%% file: main_final.tex
\documentclass[10pt,twocolumn,letterpaper]{article}

\usepackage{iccv}              


\definecolor{iccvblue}{rgb}{0.21,0.49,0.74}
\usepackage[pagebackref,breaklinks,colorlinks,allcolors=iccvblue]{hyperref}

\usepackage{colortbl} 
\usepackage{booktabs}
\usepackage{multirow}
\usepackage{graphicx}
\usepackage[table]{xcolor}
\usepackage{pgfplots} 

\usepackage{pgfplots}
\pgfplotsset{compat=1.18}  

\def\paperID{11768} 
\def\confName{ICCV}
\def\confYear{2025}

\title{MAESTRO: Task-Relevant Optimization via Adaptive Feature Enhancement and Suppression for Multi-task 3D Perception}

\author{
Changwon Kang\textsuperscript{1\thanks{Equal contribution.}}
\and
Jisong Kim\textsuperscript{1*}
\and
Hongjae Shin\textsuperscript{2*}
\and
Junseo Park\textsuperscript{2}
\and
Jun Won Choi\textsuperscript{2\thanks{Corresponding author.}}
\and
\textsuperscript{1}Hanyang University and \textsuperscript{2}Seoul National University
\\
{\tt\small \{changwonkang, jskim\}@spa.hanyang.ac.kr, \{hjshin, jspark\}@spa.snu.ac.kr, junwchoi@snu.ac.kr}
}

\begin{document}
\maketitle

\begin{abstract}
The goal of multi-task learning is to learn to conduct multiple tasks simultaneously based on a shared data representation. While this approach can improve learning efficiency, it may also cause performance degradation due to task conflicts that arise when optimizing the model for different objectives.
To address this challenge, we introduce MAESTRO, a structured framework designed to generate task-specific features and mitigate feature interference in multi-task 3D perception, including 3D object detection, bird's-eye view (BEV) map segmentation, and 3D occupancy prediction.
MAESTRO comprises three components: the Class-wise Prototype Generator (CPG), the Task-Specific Feature Generator (TSFG), and the Scene Prototype Aggregator (SPA).
CPG groups class categories into foreground and background groups and generates group-wise prototypes. The foreground and background prototypes are assigned to the 3D object detection task and the map segmentation task, respectively, while both are assigned to the 3D occupancy prediction task. TSFG leverages these prototype groups to retain task-relevant features while suppressing irrelevant features, thereby enhancing the performance for each task. SPA enhances the prototype groups assigned for 3D occupancy prediction by utilizing the information produced by the 3D object detection head and the map segmentation head. Extensive experiments on the nuScenes and Occ3D benchmarks demonstrate that MAESTRO consistently outperforms existing methods across 3D object detection, BEV map segmentation, and 3D occupancy prediction tasks.

\end{abstract}

\section{Introduction}

\begin{figure}[t]
    \centering
    \includegraphics[scale=1.3]{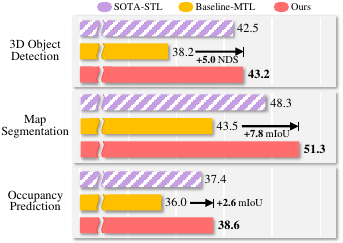} 
    \caption{
    \textbf{Performance comparison across three perception tasks.} MAESTRO consistently outperforms both the Baseline multi-task-learning (Baseline-MTL) model and the state-of-the-art (SOTA) single-task-learning (SOTA-STL) models across all tasks.}
    \label{fig1_graph}
\end{figure}

Accurate and computationally-efficient perception is essential for safe autonomous driving, as autonomous vehicles must continuously understand their surroundings to enable reliable planning. To capture different aspects of the driving environment, various perception tasks have been explored. For example, 3D object detection models have been developed to identify and localize dynamic objects \cite{BEVFormer, bevformerv2, polarformer, heightformer, frustumformer, BEVDet, bevdepth, sa-bev, dual-bev}.
Bird’s-Eye View (BEV) map segmentation models provide road-level information—such as lanes, road boundaries, and crosswalks—in the BEV  domain \cite{CVT, bevsegformer, m2bev, bevfusion, LSS, bev-seg}, offering crucial guidance on safely drivable areas.
3D occupancy prediction models discretize the environment into 3D voxels, assigning each voxel an occupancy state and semantic label to construct a dense and structured 3D representation of the scene \cite{Monoscene, fbocc, FastOcc, Surroundocc, Occ3d, OccFormer}.
As a unified perception system, these models must be executed simultaneously and in real time to support safe and effective planning in autonomous driving.

\begin{figure*}[t!]
    \centering
    \includegraphics[scale=0.54]{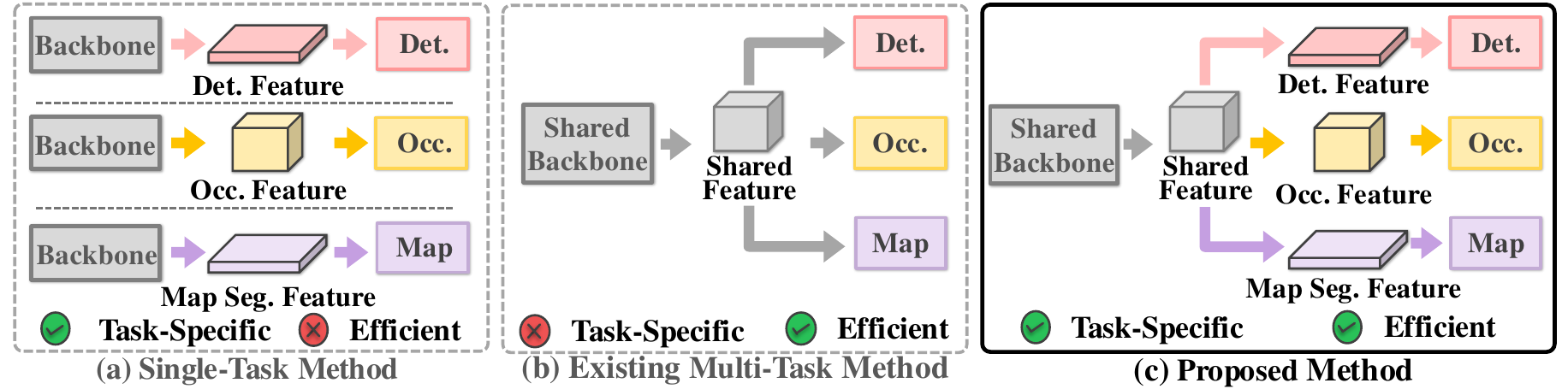}
    \caption{
    \textbf{Comparison of Single-Task and Multi-Task 3D Perception Frameworks.} (a) Single-task method: Each task utilizes a dedicated backbone and task-specific features, ensuring task specialization at the expense of high computational cost. (b) Existing multi-task method: A shared backbone improves computational efficiency but lacks task-specific feature representation. (c) Proposed method: Task-specific features are generated from shared backbone features, achieving both efficiency and task-aware feature learning.  
    }
    \label{comparison_mtl}
\end{figure*}

While these perception models can be designed and optimized independently, training and deploying separate models for each task, as shown in Figure \ref{comparison_mtl} (a), results in inefficiencies in both computation and data utilization. Since redundancy exists across tasks, it is reasonable to share a part of the architecture to exploit their commonalities.
Multi-task learning (MTL) has been investigated as a means to design architectures that share common representations while also learning task-specific information across multiple tasks.
In addition to reducing model complexity, MTL enables more efficient use of datasets by jointly training all tasks. This is particularly important for 3D perception tasks, which demand real-time performance and efficient resource utilization.

A straightforward approach to MTL is to share a backbone network that extracts features in the 3D domain for all tasks, as illustrated in Figure \ref{comparison_mtl} (b).
However, this naive approach often suffers from performance degradation due to task conflicts that arise when optimizing the shared backbone with different task-specific loss functions. The gradient signals from each loss may not be well-aligned, which limits the performance of individual tasks.

In 3D perception, different tasks focus on different semantic cues and spatial regions. For instance, 3D object detection models prioritize movable foreground objects, while BEV map segmentation models focus on static background structures. 3D occupancy estimation models, in contrast, may require attention to both foreground and background elements. These divergent behaviors can lead to conflicting supervisory signals, limiting the performance of MTL.

While various MTL methods have been proposed to address these problems in \cite{taa, sogdet, lidarformer, repvf, maskbev}, they still exhibit inferior performance compared to specialized single-task approaches \cite{bevfusion, diffuser, sogdet, panoocc, metabev, petrv2}. This persistent gap underscores the need for an effective MTL approach specifically tailored for 3D perception.

In this paper, we introduce MAESTRO (Multi-task Adaptive Feature Enhancement and Suppression for Task-Relevant Optimization), a framework designed to jointly optimize the performance of 3D perception tasks.
As illustrated in Figure \ref{comparison_mtl} (c), our method transforms shared backbone features into task-specific representations, explicitly designed to minimize interference between tasks. This improves the performance of our model for all tasks. 

MAESTRO consists of three key modules: the Class-wise Prototype Generator (CPG), the Task-Specific Feature Generator (TSFG), and the Scene Prototype Aggregator (SPA).
CPG partitions the semantic categories into foreground and background groups and generates corresponding group-wise feature prototypes~\cite{prototype}. The foreground prototypes are allocated to the 3D object detection task, while the background prototypes are used for BEV map segmentation. Both sets of prototypes are utilized in the 3D occupancy prediction task.
The resulting prototypes serve as semantic priors to guide the generation of task-specific features. 
TSFG then generates task-specific features by adaptively transforming the shared backbone features to suit each task. It effectively emphasizes task-relevant information through a feature enhancement module, while suppressing irrelevant signals via a feature suppression module. By filtering out interfering components, this mechanism helps mitigate task conflicts and improve overall multi-task learning performance.
SPA is specifically designed to enhance the performance of the 3D occupancy prediction task. It is motivated by the observation that information produced by 3D object detection and BEV map segmentation can be leveraged to benefit occupancy prediction. SPA integrates the prototypes obtained from the object detection and map segmentation heads into the prototypes assigned to the occupancy prediction module. This integration enriches the query features used for 3D occupancy prediction without compromising the performance of the other tasks.

We evaluate MAESTRO on the nuScenes benchmark \cite{nuscenes}. As shown in Figure \ref{fig1_graph}, our experiments demonstrate that MAESTRO achieves substantial performance improvements over the baseline MTL method. Remarkably, it even outperforms independently trained single-task models. This performance gain is attributed to MAESTRO’s ability to effectively leverage shared structures across tasks while minimizing inter-task interference. Moreover, MAESTRO  significantly outperforms existing MTL methods for all three 3D perception tasks.

The main contributions of this work are summarized as follows:

\begin{itemize}

\item We propose MAESTRO, a novel multi-task learning (MTL) framework that effectively generates task-specific features while mitigating feature interference across multiple 3D perception tasks.

\item We introduce a group-wise prototype generation method that clusters semantic classes and derives representative prototypes for each group. These prototypes are used to enhance task-relevant features and suppress interfering components from the shared backbone, thereby reducing task conflicts and improving overall performance.

\item We exploit task dependencies for further performance improvement by leveraging the outputs from 3D object detection and BEV map segmentation models to enhance the performance of the 3D occupancy prediction.

\item The source code will be made publicly available to facilitate future research and reproducibility.

\end{itemize}

\begin{figure*}[t!]
    \centering
    \includegraphics[scale=0.51]{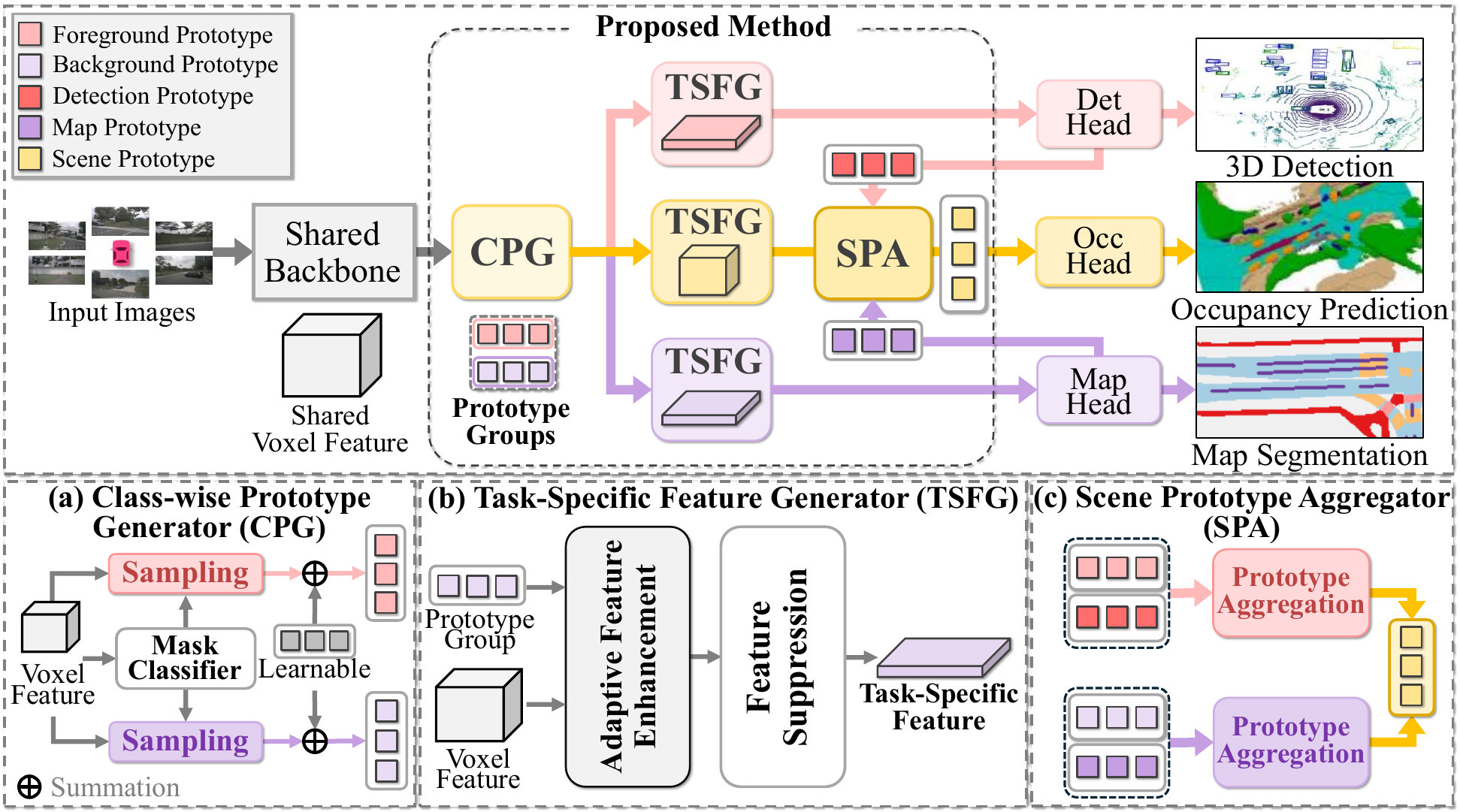}
    \caption{
    \textbf{Overall architecture of MAESTRO.} 
    Multi-view images are processed by a shared backbone to generate a structured 3D voxel representation. The CPG generates foreground and background prototype groups, which guide the TSFG in refining task-relevant features for 3D object detection, BEV map segmentation, and 3D occupancy prediction. Task-oriented prototypes derived from the 3D object detection and BEV map segmentation heads are then integrated with the prototype groups via the SPA, forming Scene Prototypes. These Scene prototypes are subsequently processed by the occupancy decoder to produce the final 3D occupancy predictions.
    }
    \label{overall}
\end{figure*}

\section{Related Works}

\subsection{Camera-based Perception Tasks}
\noindent \textbf{3D object detection.} 
Recent advances in multi-camera 3D object detection have centered around the transformation of image features into the BEV space, primarily following either query-based or depth-based projection approaches. Query-based methods generate learnable BEV queries and extract features from multi-view images via cross-attention mechanisms \cite{BEVFormer, bevformerv2, polarformer, heightformer, frustumformer}. In contrast, depth-based methods—initially introduced by LSS \cite{LSS}—explicitly project image features into BEV space by estimating depth distributions \cite{BEVDet, bevdepth, sa-bev, dual-bev}. More recent works such as SA-BEV \cite{sa-bev} and DualBEV \cite{dual-bev} improve upon this strategy by incorporating object-centric pooling, which enhances the relevance of extracted features while effectively suppressing background noise.

\noindent \textbf{BEV map segmentation.}
BEV map segmentation methods adopt BEV transformation strategies. BEVSegFormer \cite{bevsegformer} employs deformable cross-attention to project image features into BEV space without requiring depth estimation or camera parameters. In contrast, explicit warping methods \cite{m2bev, bevfusion, LSS, bev-seg} rely on camera intrinsics and estimated depth to perform geometric projections. M2BEV \cite{m2bev} further refines this approach by aligning features to the ego-vehicle frame, while BEV-Seg \cite{bev-seg} improves projection quality through a dedicated parsing network and enhanced depth estimation.

\noindent \textbf{3D occupancy prediction.} 
MonoScene \cite{Monoscene} pioneered 3D occupancy prediction using a single image and a 3D U-Net architecture. BEV-based methods \cite{fbocc, FastOcc, flashocc} simplify computation by collapsing the height dimension, while voxel-based approaches \cite{Surroundocc, Occ3d, protoocc} adopt a coarse-to-fine strategy to produce dense and structured 3D predictions. TPVFormer \cite{TPVFormer} introduces a tri-perspective view representation, but its performance is limited by the loss of detailed 3D information. In contrast, OccFormer \cite{OccFormer} advances voxel-based semantic occupancy prediction through a dual-path transformer architecture.

\subsection{Multi-Task Learning}
MTL methods are broadly categorized into optimization-based and architecture-based approaches. Optimization-based methods aim to balance task contributions through techniques such as loss weighting and gradient manipulation \cite{uncertaintyloss, gradnorm, just, conflict, dynamicmtl, multiobj}. Architecture-based methods can be further divided into encoder-focused and decoder-focused designs. Encoder-focused approaches \cite{cross_stitch, mtl_nas, latent} emphasize learning shared representations across tasks, while decoder-focused approaches \cite{padnet, tcdcn, mrn_mtl} utilize a shared encoder followed by task-specific decoders to promote task-aware predictions. Recent decoder-focused models, such as InvPT \cite{invpt} and InvPT++ \cite{invptv2}, employ transformer-based decoders for enhanced task-specific feature extraction. Similarly, TaskExpert \cite{taskexpert} leverages a Mixture-of-Experts (MoE) framework with spatial context-aware gating to dynamically route features to task-relevant experts.

In the context of 3D perception, MTL methods predominantly adopt decoder-focused architectures \cite{sogdet, bevfusion, metabev}, utilizing a shared encoder with multiple decoders to address various scene understanding tasks. However, empirical findings from BEVFormer \cite{BEVFormer} and M2BEV \cite{m2bev} reveal that naive joint training often leads to degraded task performance due to feature interference. To address this, SOGDet \cite{sogdet} introduces a weighted feature-sharing mechanism for effective modality fusion, while HENet \cite{henet} dynamically adjusts feature resolutions to better align with task requirements. Although these methods alleviate task conflict and improve overall performance, they still struggle to fully capture task-specific feature representations.


\section{Proposed Methods}
\subsection{Overview}
The overall structure of MAESTRO is illustrated in Figure~\ref{overall}.
2D feature maps are independently extracted from multi-view camera images and subsequently lifted into the 3D domain using the Lift-Splat-Shoot (LSS) method \cite{LSS}, resulting in a voxel representation $F_{\text{s}}\in \mathbb{R}^{C \times X \times Y \times Z}$ for downstream tasks. These voxel features serve as shared backbone features across all tasks.
The CPG first groups class categories into foreground and background sets. It then applies a mask classifier to produce a binary voxel mask for each class. Based on these masks, CPG generates group-wise prototype features by sampling and pooling masked voxel features from \( F_{\text{s}} \).
These prototype groups are assigned to their respective tasks, ensuring that each group is semantically aligned with its corresponding task.
Next, the TSFG refines the shared backbone features using the task-assigned prototype groups via Adaptive Feature Enhancement and Feature Suppression modules. This process yields task-specific features that are optimized for each perception task.
These features are subsequently passed through the 3D object detection, BEV map segmentation, and 3D occupancy prediction heads.
The SPA further refines the prototype group associated with 3D occupancy prediction by incorporating outputs from the 3D object detection and BEV map segmentation heads. 
Finally, these adapted prototypes are then used to initialize the queries for the 3D occupancy decoder.

\subsection{Class-wise Prototype Generator (CPG)}
Naive feature sharing in MTL often results in representations that lack clear semantic separation across tasks. To address this limitation, we introduce the CPG whose structure is depicted in Figure~\ref{overall} (a). CPG first organizes semantic categories into foreground and background groups. For example, the foreground group includes object classes such as car, truck, pedestrian, and bicycle, while the background group comprises semantic categories like drivable surface, sidewalk, manmade structures, and vegetation.
The CPG then generates prototype features for each group. 
Let $\mathcal{K}=\{1,\dots, K\}$ denote the set of all $K$ semantic categories. CPG groups them into foreground and background groups $\mathcal{K}_{fg}$ and $\mathcal{K}_{bg}$. 
CPG applies a lightweight mask classifier to  $F_{\text{s}}$ to compute voxel-wise semantic confidence scores $S_v \in \mathbb{R}^{K \times X \times Y \times Z}$. The most confident class is assigned to each voxel, forming class-wise masks $B_k \in \mathbb{R}^{X \times Y \times Z}$ as
\begin{equation}
B_k(i,j,l) = \mathbb{I} \left( \underset{k' \in \{1, \dots, K\}}{\mathrm{argmax}} \, S_v(i,j,l) = k \right),
\end{equation}
where $(i, j, l)$ represents voxel coordinates, and \( \mathbb{I}(\cdot) \) is an indicator function that returns \( 1 \) if the voxel is assigned to class \(k\) and \( 0 \) otherwise.
The prototype \( P_k \in \mathbb{R}^{C} \) for the $k$-th class is obtained by applying average pooling to the masked features as $P_k = \text{AvgPool}(F_{s} \otimes B_{k})$, where \( \otimes \) denotes element-wise multiplication. If no voxels belong to class $k$, $P_k$ is set to zero for consistency.
The group-wise prototypes are generated by gathering the prototypes for each group and adding learnable embeddings, i.e., $P_{FG}=\{P_k + E_{k}\}_{k \in \mathcal{K}_{fg}} \in \mathbb{R}^{N_{fg} \times C}$ and $P_{BG}=\{P_k + E_{k}\}_{k \in \mathcal{K}_{bg}} \in \mathbb{R}^{N_{bg} \times C}$, where $N_{fg}$ and $N_{bg}$ represent the number of foreground and background classes, and $E_k$ is a learnable embedding.
Finally, the prototype groups $P_{FG}$ and $P_{BG}$ are assigned to the appropriate task. Considering the objective of 3D perception tasks, the group assignment can be determined as  $G_{\text{Det}} = P_{\text{FG}}$, $G_{\text{Map}} = P_{\text{BG}}$, and $G_{\text{Occ}} = P_{\text{FG}} \cup P_{\text{BG}}$, where  $G_{\text{Det}}$, $G_{\text{Map}}$, and $ G_{\text{Occ}}$ are the prototype groups assigned to 3D object detection, BEV map segmentation, and 3D occupancy prediction tasks, respectively.
These task-specific prototype groups $G_{\text{Det}}$, $G_{\text{Map}}$, and $ G_{\text{Occ}}$ are used for adaptive feature enhancement in TSFG.

\subsection{Task-Specific Feature Generator (TSFG)}
\begin{figure}[t]
    \centering
    \includegraphics[scale=0.36]{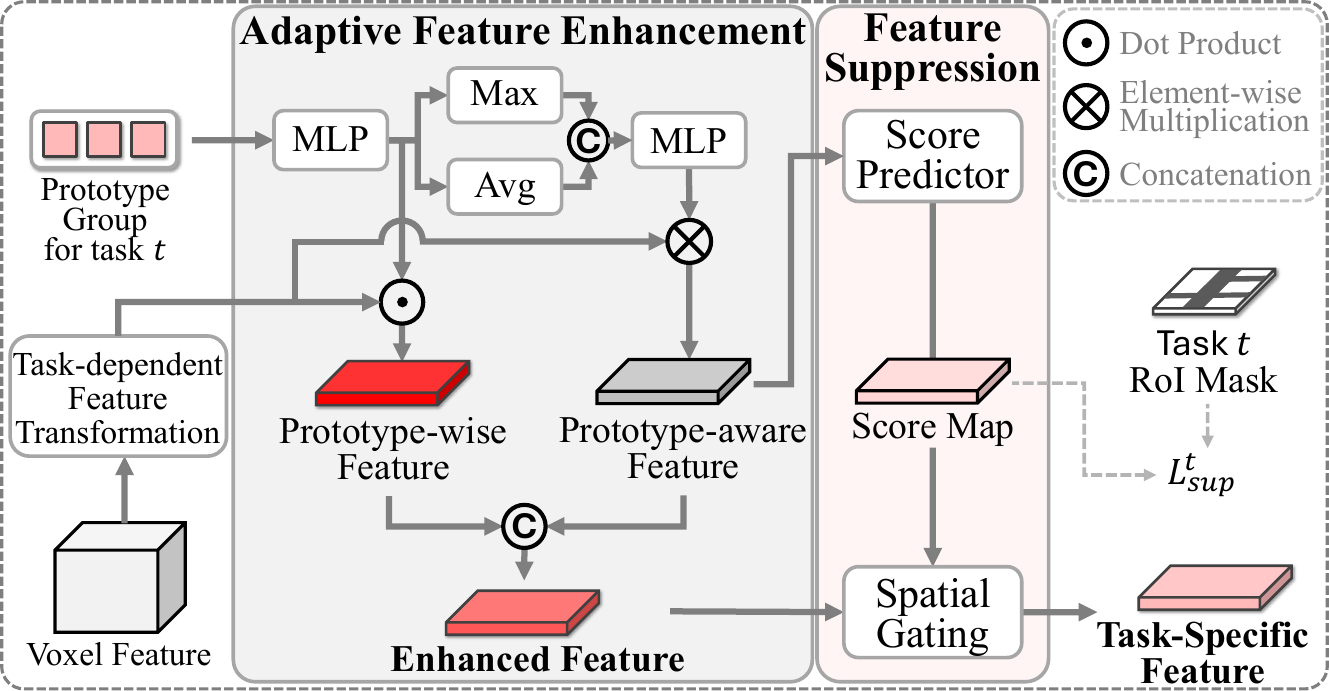}
    \caption{\textbf{Detailed structure of TSFG.} TSFG refines task-relevant features by leveraging the prototype group, followed by feature suppression to mitigate task-irrelevant information, resulting in task-specific features for each task.
    }
    \label{fig_tsfg}
\end{figure}

The structure of the TSFG is shown in Figure~\ref{fig_tsfg}. TSFG transforms the shared backbone features into task-specific features tailored to each of the three 3D perception tasks. The TSFG module operates through three sequential stages: (1) Task-dependent Feature Transformation, (2) Adaptive Feature Enhancement, and (3) Feature Suppression.

\noindent \textbf{Task-dependent Feature Transformation.}
Task-dependent Feature Transformation module first transforms the shared voxel features \( F_{\text{s}} \) into representations suited for each task. For BEV-oriented tasks, such as 3D object detection and BEV map segmentation, it generates features in the BEV domain. In contrast, for voxel-oriented tasks like 3D occupancy prediction, it produces features in the voxel domain.
Specifically, for BEV-oriented tasks, the height axis \( Z \) of \( F_{\text{s}} \) is collapsed into the channel dimension, and a series of 2D convolutional layers is applied to produce BEV features \( F^{\text{BEV}}_{t} \in \mathbb{R}^{C \times X \times Y} \), where \( t \in \{\mathrm{Det}, \mathrm{Map}\} \). For voxel-oriented tasks, \( F_{\text{s}} \) is processed with a sequence of 3D convolutional layers to generate voxel features \( F^{\text{voxel}}_{\mathrm{Occ}} \in \mathbb{R}^{C \times X \times Y \times Z} \).
These transformed features are then passed to the Adaptive Feature Enhancement module.

\noindent \textbf{Adaptive Feature Enhancement.}
Adaptive Feature Enhancement module enhances the transformed features $F^{\text{BEV}}_t$ or $F^{\text{voxel}}_t$ using the corresponding prototype group $G_t$, where $t\in\{\text{Det}, \text{Map}, \text{Occ}\}$.
After projecting the prototypes in $G_t$ through an MLP, it generates the prototype-wise features by taking a dot-product operation between each prototype group and the transformed features. Denote the prototype-wise features as $F^{\text{BEV}}_{\text{wise}} \in \mathbb{R}^{N_{G_t} \times X \times Y}$ for BEV-oriented tasks and $F^{\text{voxel}}_{\text{wise}} \in \mathbb{R}^{N_{G_t} \times X \times Y \times Z}$ for voxel-oriented tasks, where $N_{G_t}$ is the number of prototypes in $G_t$. In effect, these prototype-wise features activate spatial regions that are semantically aligned with $G_t$.
In parallel, Adaptive Feature Enhancement module generates the prototype-aware features via channel attention operation. 
First, the global context features are extracted from the prototype group $G_t$ through max and average pooling.
These pooled features are concatenated and processed by an MLP, producing a channel scaling vector $\gamma^{scale}_t\in\mathbb{R}^{C}$. This scaling vector modulates the transformed features $F^{\text{BEV}}_t$ or $F^{\text{voxel}}_t$ via element-wise multiplication, resulting in prototype-aware features $F^{\text{BEV}}_{\text{aware}}\in\mathbb{R}^{C\times X\times Y}$ or $F^{\text{voxel}}_{\text{aware}}\in\mathbb{R}^{C\times X\times Y\times Z}$.
Finally, the prototype-wise features and the prototype-aware features are concatenated and passed through convolution layers to yield the enhanced feature $\tilde{F}_t$, i.e.,  
\begin{equation}
\tilde{F}_{t} =
\begin{cases}
\operatorname{Conv2D}\bigl(\operatorname{Cat}[F^{\text{BEV}}_{\text{wise}}; F^{\text{BEV}}_{\text{aware}}]\bigr), & t\in\{\text{Det, Map}\},\\[6pt]
\operatorname{Conv3D}\bigl(\operatorname{Cat}[F^{\text{voxel}}_{\text{wise}}; F^{\text{voxel}}_{\text{aware}}]\bigr), & t=\text{Occ},
\end{cases}
\end{equation}
where $\operatorname{Cat}[\cdot;\cdot]$ indicates channel-wise concatenation.
The enhanced features and the prototype-aware features are passed to the subsequent suppression module.

\noindent \textbf{Feature Suppression.}
While the previous process emphasizes task-relevant information using guidance from a prototype group, the Feature Suppression module mitigates irrelevant components to mitigate task interference. It applies a lightweight CNN to the prototype-aware features, generating suppression scores, which ideally approach one within the Region of Interest (RoI) region and zero elsewhere. Here, the RoI region is specified by RoI masks, which indicate bounding boxes in the BEV domain for 3D object detection, semantically active areas for BEV map segmentation, and occupied voxels for 3D occupancy prediction.
The Suppression Score Map $S^{\text{supp}}_{t}$ is computed from the prototype-aware features as
\begin{equation}
S^{\text{supp}}_{t} =
\begin{cases}
f^{2D}_{score}(F^{\text{BEV}}_{\text{aware}})\in\mathbb{R}^{X\times Y}, & t\in\{\text{Det, Map}\},\\[6pt]
f^{3D}_{score}(F^{\text{voxel}}_{\text{aware}})\in\mathbb{R}^{X\times Y\times Z}, & t=\text{Occ},
\end{cases}
\end{equation}
where $f^{2D}_{score}$ and $f^{3D}_{score}$ denote the score predictors for BEV and voxel-oriented tasks, respectively. Note that the generation of the Suppression Score Map is supervised using the ground truth.
The resulting Suppression Score Map $S^{\text{supp}}_{t}$ is multiplied to the enhanced feature $\tilde{F}_t$ through a gating operation. This operation yields the task-specific features as 
$F^{\text{TS}}_{t}=\tilde{F}_t\otimes S^{\text{supp}}_t$,
where $\otimes$ represents element-wise multiplication. This design effectively filters out task-irrelevant components from the input features.
These resulting $F^{\text{TS}}_{t}$ are subsequently utilized by task-dedicated heads and the SPA.

\subsection{Scene Prototype Aggregator (SPA)}
SPA operates in two stages.
First, it generates task-oriented prototypes based on the predicted outputs from the object detection and map segmentation heads.
Then, these task-oriented prototypes are integrated into the original prototype group, resulting in the scene prototypes that serve as initial queries for the 3D occupancy decoder. (see Figure~\ref{overall} (c).) 

\noindent \textbf{Task-oriented prototype generation.}
SPA produces task-oriented prototypes using the predicted bounding boxes from the object detection head and the segmentation masks from the BEV map segmentation head.
Specifically, detection prototypes \(P_{\text{Det}} \in \mathbb{R}^{N_{\text{Det}} \times C}\) are generated by applying RoIAlign \cite{maskrcnn} to the task-specific features \(F^{\text{TS}}_{\text{Det}}\), followed by class-wise average pooling over the predicted bounding boxes, where \(N_{\text{Det}}\) denotes the number of detection classes.
In parallel, map prototypes \(P_{\text{Map}} \in \mathbb{R}^{N_{\text{Map}} \times C}\) are computed by masked average pooling over the task-specific features \(F^{\text{TS}}_{\text{Map}}\), using the predicted masks, where $N_{\text{map}}$ is the number of classes defined in the BEV map segmentation task.
Both $P_{\text{Det}}$ and $P_{\text{Map}}$ constitute task-oriented prototypes.

\noindent \textbf{Prototype aggregation.}
SPA employs explicit semantic aggregation rules to align heterogeneous semantic label spaces across tasks.
First, the prototypes belonging to categories that have a one-to-one semantic correspondence between the prototype groups $G_t$  and the task-oriented prototypes are aggregated via direct class-wise summation.
In contrast, prototypes associated with multiple fine-grained labels are first averaged and subsequently summed with their corresponding prototypes in $G_t$.
This rule-based aggregation yields scene prototypes, which are used to initialize the occupancy queries in the occupancy decoder.
By leveraging complementary semantics from 3D object detection and BEV map segmentation tasks, SPA enhances occupancy prediction performance while preserving the effectiveness of the other tasks.

\begin{table*}[t]
\centering
\fontsize{9pt}{9pt}\selectfont  
\setlength{\tabcolsep}{5pt}
\begin{tabular}{l|c|c|c|cc|c|c|c}
\toprule[1.2pt]
\multicolumn{1}{c|}{\textbf{Method}} & \textbf{Venue}  & \textbf{Image Backbone} & \textbf{Image Size} & \textbf{mAP} & \textbf{NDS}  & \textbf{mIoU (Map)} & \textbf{mIoU (Occ)} & \textbf{Latency (ms)} \\ 
\midrule[0.4pt]
BEVDet \cite{BEVDet}            & arXiv'21 & ResNet-50 & $256\times704$         & 29.8  & 37.9  & - & -  &   51.5         \\
DETR3D \cite{detr3d}            & CoRL'22  & ResNet-50 & $800\times1333$         & 30.3  & 37.4  & - & -  &   -         \\
BEVDepth * \cite{bevdepth}        & AAAI'23  & ResNet-50 & $256\times704$        & 33.7  & 41.4  & - & -  & -           \\
BEVFormer v2 \cite{bevformerv2} & CVPR'23 & ResNet-50 & $640\times1600$          & 35.1  & 41.4  & - & -  & 171.9         \\
DualBEV $\dagger$  \cite{dual-bev}       & ECCV'24 & ResNet-50 & $256\times704$          & 35.2  & 42.5  & - & -  & 65.1           \\
\midrule[0.4pt]
CVT \cite{CVT}                  & CVPR'22 & ResNet-50          & $256\times704$ & -     & -  & 37.7 & -  & 33.9         \\
LSS \cite{LSS}                  & ECCV'20 & ResNet-50          & $256\times704$ & -     & -  & 41.0 & -   & 72.4        \\
BEVFusion \cite{bevfusion}      & ICRA'23 & ResNet-50          & $256\times704$ & -     & -  & 47.1 & -  & 45.6      \\
DifFUSER  \cite{diffuser}       & ECCV'24 & ResNet-50          & $256\times704$ & -     & -  & 48.3 & -  & 92.2      \\
\midrule[0.4pt]
MonoScene \cite{Monoscene}      & CVPR'22  & ResNet-101 & $928\times1600$          & -     & -  & -  & 6.1 & 830.1           \\
OccFormer \cite{OccFormer}      & ICCV'23  & ResNet-50  & $928\times1600$          & -     & -  & -  & 21.9  & 349.0           \\
TPVFormer \cite{TPVFormer}      & CVPR'23  & ResNet-101 & $928\times1600$          & -     & -  & -  & 27.8  & 320.8           \\
Vampire \cite{Vampire}          & AAAI'24  & ResNet-101 & $256\times704$           & -     & -  & -  & 28.3   & 349.2          \\
CTF-Occ \cite{Occ3d}            & NIPS'24  & ResNet-101 & $928\times1600$          & -     & -  & -  & 28.5  & -           \\
SurroundOcc \cite{Surroundocc}  & ICCV'23  & ResNet-101 & $800\times1333$          & -     & -  & -  & 34.6  & 355.6           \\
FB-Occ \cite{fbocc}             & ICCV'23  & ResNet-50 & $256\times704$           & -     & -  & -  & 37.4  & 129.7           \\
\midrule[0.4pt]
BEVFusion \cite{bevfusion}      & ICRA'23 & ResNet-50          & $256\times704$ & 33.6  & 39.2 & 44.0 & - & - \\
PanoOcc \cite{panoocc}          & CVPR'24 & ResNet-50          & $864\times1600$& 29.5  & 34.8 & - & 31.8   & 124.7 \\
\midrule[0.4pt]
\rowcolor[gray]{0.90}
\multicolumn{1}{c|}{Baseline -STL}                    & - & ResNet-50          & $256\times704$ & 33.8  & 41.7  & 47.5 & 36.5  & 405.9 \\
\rowcolor[gray]{0.90}
\multicolumn{1}{c|}{Baseline-MTL} & - & ResNet-50 & $256\times704$ & 32.7 & 38.2 & 43.5 & 36.0   & 219.6        \\
\rowcolor[gray]{0.90}
\multicolumn{1}{c|}{Ours-MTL} & - & ResNet-50 & $256\times704$ & \textbf{36.4} & \textbf{43.2} & \textbf{51.3} & \textbf{38.6}   & 250.3   \\
\bottomrule[1.2pt]
\end{tabular}
\caption{
Performance comparison evaluated on the nuScenes validation set.
``-" indicates tasks not supported by the method. All BEV map segmentation and MTL results were reproduced using official code with ResNet-50 backbone. ``*" denotes the performance reported by DualBEV \cite{dual-bev}. $\dagger$ indicates the method using CBGS for training. Our MTL approach achieves state-of-the-art performance across all tasks.
}
\label{single_frame}
\end{table*}

\subsection{Training Loss}
The total loss function $L_{total}$ is given by
\begin{equation}
    L_{total} = L_{depth} + L_{CPG} + L_{Sup} + L_{det} + L_{map} + L_{occ} ,
\end{equation}
where $L_{depth}$ is for depth estimation, $L_{CPG}$ is for class-wise mask classification in CPG, $L_{Sup}$ is the suppression loss summed across tasks used for supervising suppression score maps, and $L_{det}$, $L_{map}$, and $L_{occ}$ correspond to the task-specific losses for 3D object detection, BEV map segmentation, and 3D occupancy prediction, respectively.
Dice and Lovasz losses \cite{lovasz} are used in $L_{CPG}$ while Focal loss \cite{focalloss} is utilized in $L_{Sup}$.
The detection loss \(L_{det}\) is formulated based on CenterPoint \cite{centerpoint} for object detection. The map segmentation loss \(L_{map}\) is adopted from BEVFusion \cite{bevfusion}, and the occupancy prediction loss \(L_{occ}\) is based on the loss function employed in OccFormer \cite{OccFormer}.

\section{Experiments}
\subsection{Experimental Settings}
\noindent \textbf{Datasets and metrics.} 
Our experiments were conducted on the nuScenes \cite{nuscenes} dataset. We particularly used the annotation provided in Occ3D \cite{Occ3d} for 3D  occupancy prediction. 
For 3D object detection evaluation, we adopt the official metrics: mean Average Precision (mAP) and nuScenes Detection Score (NDS). BEV map segmentation performance is evaluated following \cite{bevfusion, CVT}, using mean Intersection-over-Union (mIoU) across six predefined background classes. The 3D semantic occupancy prediction performance is evaluated using voxel-level mIoU across 17 semantic categories.
The details of the dataset and the performance metrics are provided in the Supplementary Material.

\noindent \textbf{Implementation details.}
We utilized ResNet-50 \cite{ResNet} for the image backbone. Our model was trained for 24 epochs, without employing class-balanced grouping and sampling (CBGS) \cite{cbgs}. The AdamW optimizer is used with a learning rate of $1 \times 10^{-4}$ and weight decay of 0.01, with a batch size of 8 distributed across 4 RTX 3090 GPUs. Baseline-STL refers to models independently optimized for each task, while Baseline-MTL denotes a naive multi-task learning approach without the proposed modules, as illustrated in Figure~\ref{comparison_mtl} (b). We also compare the performance of MAESTRO with those of existing STL and MTL methods. We include BEVFusion \cite{bevfusion} and PanoOcc \cite{panoocc} as MTL baselines, as they are the officially available multi-task methods on the nuScenes benchmark. Further training details are provided in the Supplementary Material.

\begin{table}[t!]
    \centering
    \setlength{\tabcolsep}{2pt}
    \renewcommand{\arraystretch}{1.1} 
    \fontsize{9pt}{9pt}\selectfont
    \begin{tabular}{>{\centering\arraybackslash}m{0.7cm}|>{\centering\arraybackslash}m{0.6cm}|>{\centering\arraybackslash}m{0.6cm}|>{\centering\arraybackslash}m{0.6cm}|c|cc|c|c}
    \toprule[1.2pt]
        \multirow{2}{*}{\textbf{CPG}} & \multicolumn{3}{c|}{\textbf{TSFG}} & \multirow{2}{*}{\textbf{SPA}} & \multirow{2}{*}{\textbf{mAP}} & \multirow{2}{*}{\textbf{NDS}} & \multirow{1.5}{*}{\textbf{mIoU}} & \multirow{1.5}{*}{\textbf{mIoU}} \\
        \cline{2-4}
        &Det. & Map. & Occ. & & & & \textbf{(Map)}&\textbf{(Occ)} \\
        \midrule[0.4pt]
                     &            &            &            &            & 31.3 & 32.3 & 40.4 & 33.6 \\  
        \checkmark   &            &            &            &            & 31.3 & 32.4 & 40.3 & 34.6 \\  
        \checkmark   &\checkmark  &            &            &            & 32.6 & 34.1 & 39.1 & 34.1 \\  
        \checkmark   &            & \checkmark &            &            & 30.8 & 31.9 & 44.0 & 34.3 \\  
        \checkmark   &            &            & \checkmark &            & 31.1 & 32.2 & 41.2 & 35.9 \\  
        \checkmark   &\checkmark  & \checkmark & \checkmark &            & 32.5 & 34.1 & 44.1 & 35.9 \\  
        \midrule[0.4pt]
        \rowcolor[gray]{0.90}
        \checkmark &\checkmark  & \checkmark & \checkmark & \checkmark & \textbf{32.6} & \textbf{34.3} & \textbf{44.2} & \textbf{36.9} \\  
        \bottomrule[1.2pt]
    \end{tabular}
    \caption{Ablation study for evaluating the main components in MAESTRO.}
    \label{main_Abl}
\end{table}


\subsection{Comparison with State-of-the-art Results}
Table~\ref{single_frame} presents the performance of MAESTRO on the nuScenes validation set. We observe that the Baseline-MTL model performs worse than the Baseline-STL model due to task conflicts inherent in naive MTL. In contrast, MAESTRO, which incorporates the proposed techniques, demonstrates substantial performance improvements over Baseline-STL across all three tasks: a 2.6\% increase in mAP for 3D object detection, a 3.8\% increase in mIoU for BEV map segmentation, and a 2.1\% increase in mIoU for 3D occupancy prediction. The improvements are even more pronounced when compared to Baseline-MTL. 
MAESTRO significantly reduces latency by 155.6 ms compared to Baseline-STL, with only a marginal increase of 30.7 ms over Baseline-MTL.
We also compare MAESTRO against existing STL and MTL methods. Notably, MAESTRO significantly outperforms the previous best-performing STL models for each individual task. Furthermore, compared to existing MTL approaches, MAESTRO achieves state-of-the-art performance.


\subsection{Ablation Studies}
We conducted ablation studies to evaluate the effectiveness of the proposed ideas. Our ablation study was conducted on 1/4 of the nuScenes training set for 24 epochs with all other settings unchanged.

\noindent \textbf{Component analysis.}
Table~\ref{main_Abl} presents the performance of the proposed method as each component is incrementally added. The baseline model adopts a naive MTL approach that shares features across all tasks without any of the proposed enhancements, as illustrated in Figure~\ref{comparison_mtl}(b). Integrating the CPG into the baseline yields a 1.0\% mIoU improvement in occupancy prediction without impacting the performance for other tasks. This demonstrates the benefit of incorporating semantic prototypes. Applying the TSFG to a single task enhances the performance of that task but results in performance degradation in others due to task imbalance. When TSFG is applied to all tasks, it effectively mitigates task interference, leading to improvements of 1.2\% mAP in 3D object detection, 3.8\% mIoU in BEV map segmentation, and 1.3\% mIoU in 3D occupancy prediction. Finally, the inclusion of the SPA further boosts occupancy prediction by an additional 1.0\% mIoU without compromising the performance of the other tasks.

\begin{table}[t!]
\centering
\resizebox{1.01\linewidth}{!}{
\setlength{\tabcolsep}{2.0pt}
\fontsize{9pt}{9pt}\selectfont
\begin{tabular}{l|c c|c|c}
\toprule[1.2pt]
\multicolumn{1}{c|}{\textbf{Method}} & \textbf{mAP} &\textbf{NDS} & \textbf{mIoU (Map)} & \textbf{mIoU (Occ)} \\ 
\midrule[0.4pt]
\ Baseline-MTL with CPG      & 31.3   & 32.4   & 40.3    & 34.6  \\
\ + Prototype-Wise Features   & 31.9   & 32.7   & 42.1    & 35.2 \\
\ + Prototype-Aware Features  & 32.2   & 33.6   & 43.4    & 35.7  \\
\ + Feature Suppression  & \textbf{32.5} & \textbf{34.1}  & \textbf{44.1} & \textbf{35.9} \\
\bottomrule[1.2pt] 
\end{tabular}
}
\caption{Ablation study for evaluating the components of TSFG.}
\label{TSFG_Abl}
\end{table}



\begin{table}[t!]
\centering
\fontsize{9pt}{9pt}\selectfont
\setlength{\tabcolsep}{10.5pt}
\begin{tabular}{l|c}
\toprule[1.2pt]
\multicolumn{1}{c|}{\textbf{Method}} & \textbf{mIoU (Occ)} \\ 
\midrule[0.4pt]
\ MAESTRO                      &\textbf{36.9} \\
\ w/o Map Prototypes           & 36.3 (-0.6)  \\
\ w/o Det Prototypes           & 35.9 (-1.0) \\
\bottomrule[1.2pt] 
\end{tabular}
\caption{Ablation study for evaluating components of prototype aggregation used in SPA.}
\label{Ablation_SPA}
\end{table}

\begin{figure*}[t!]
    \centering
    \includegraphics[scale=0.43]{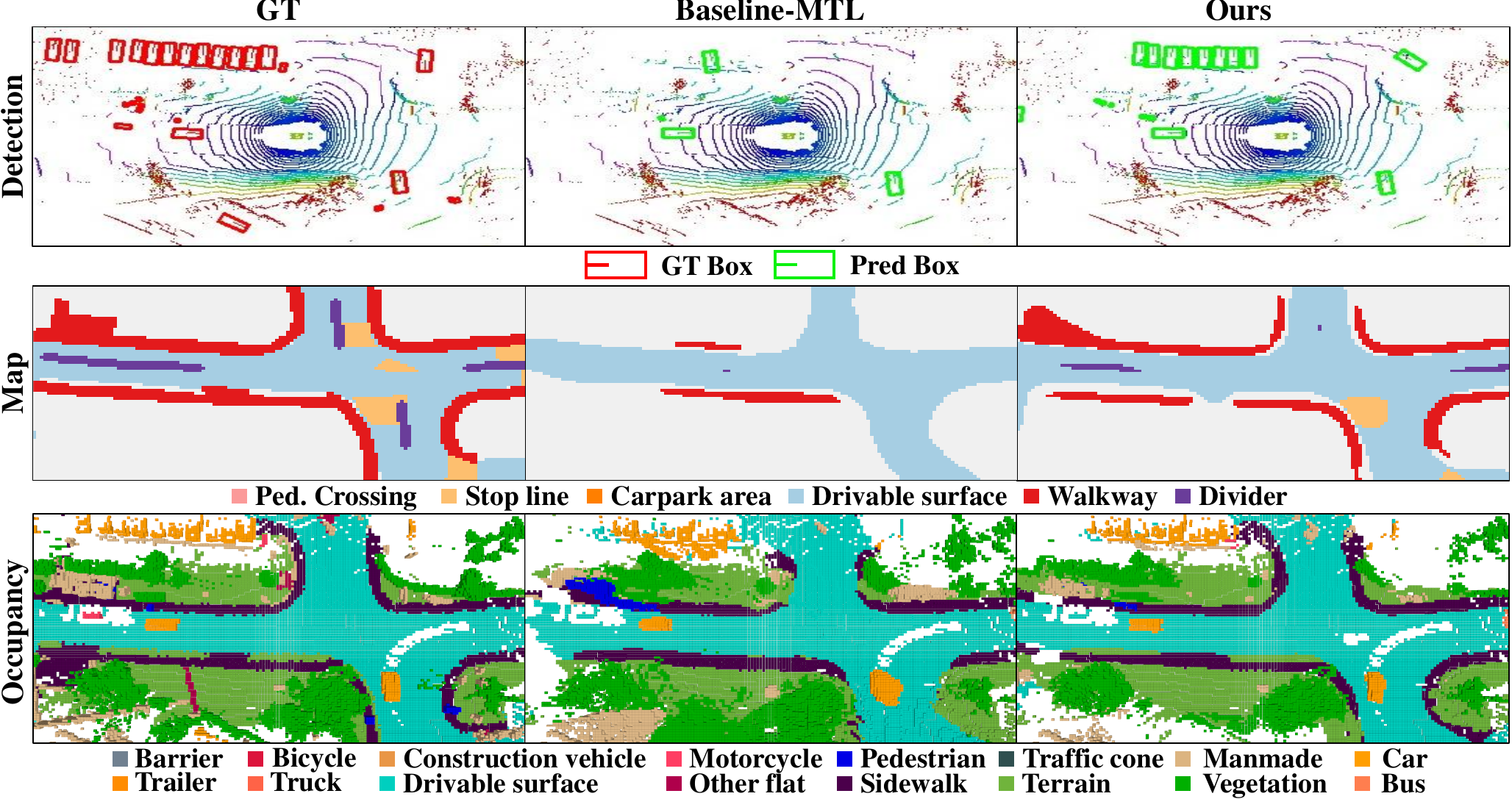}
    \caption{
    \textbf{Qualitative results on the nuScenes validation set.}
    From left to right, we show the qualitative results of Ground Truth, Baseline-MTL, and our MAESTRO.
    }
    \label{main_vis}
\end{figure*}



\noindent \textbf{Impact of adaptive feature enhancement and suppression in TSFG.}
Table~\ref{TSFG_Abl} presents an ablation study analyzing the impact of each component within TSFG.
Starting from the Baseline-MTL with CPG, we progressively add the key components of TSFG to evaluate their individual contributions. Adding the Prototype-Wise Features results in a performance gain of 0.6\% mAP in 3D object detection, 1.8\% mIoU in BEV map segmentation, and 0.6\% mIoU in 3D occupancy prediction. Incorporating the Prototype-Aware Features leads to an additional gain of 0.3\% mAP, 1.3\% mIoU, and 0.5\% mIoU for the respective tasks. Finally, adding the Feature Suppression module further improves performance, with a 0.3\% mAP gain in 3D object detection, and 0.7\% and 0.2\% mIoU gains in BEV map segmentation and 3D occupancy prediction, respectively.

\noindent \textbf{Effect of task-oriented prototypes for prototype aggregation in SPA.}
Table~\ref{Ablation_SPA} demonstrates the impact of task-oriented prototypes by progressively removing each component from the full MAESTRO model. Removing the map prototypes leads to a 0.6\% decrease in 3D occupancy prediction mIoU, highlighting their role in capturing background semantics. Further excluding detection prototypes causes an additional drop of 0.4\% mIoU, emphasizing their importance for foreground representation. These results confirm that jointly aggregating both types of prototypes significantly enhances 3D occupancy prediction performance.


\subsection{Qualitative Results}
Figure~\ref{main_vis} presents a qualitative comparison between Baseline-MTL, our method, and the ground truth across the three perception tasks. As shown, our approach yields more accurate predictions across all tasks compared to Baseline-MTL. Additional qualitative results are provided in the Supplementary Material.

\section{Conclusions}
In this paper, we presented MAESTRO, a unified framework for multi-task 3D perception that explicitly mitigates feature interference while enhancing task efficiency. MAESTRO introduces the CPG, which groups semantic categories into foreground and background sets and generates group-wise prototypes that serve as semantic priors for task-specific feature enhancement. The TSFG leverages these priors to selectively enhance task-relevant features while suppressing irrelevant regions, enabling effective and discriminative feature learning for each task. Additionally, the SPA enriches the prototype groups used for 3D occupancy prediction by integrating complementary information from the 3D object detection and map segmentation heads. This leads to improved occupancy performance without degrading the accuracy of the other tasks. Extensive evaluations on the nuScenes and Occ3D benchmarks demonstrate that MAESTRO achieves state-of-the-art performance across object detection, BEV map segmentation, and occupancy prediction tasks, validating the effectiveness and robustness of our structured multi-task learning approach.

{
    \small
    \bibliographystyle{ieeenat_fullname}
    \bibliography{main}
}

\end{document}